\documentclass[conference]{IEEEtran}
\IEEEoverridecommandlockouts

\usepackage{cite}
\usepackage{amsmath,amssymb,amsfonts}
\usepackage{graphicx}
\usepackage{textcomp}
\usepackage{xcolor}
\usepackage{booktabs}
\usepackage{array}
\usepackage{multirow}
\usepackage{hyperref}

\def\BibTeX{{\rm B\kern-.05em{\sc i\kern-.025em b}\kern-.08em
T\kern-.1667em\lower.7ex\hbox{E}\kern-.125emX}}

\begin{document}

\title{Drug-Target Interaction Prediction via Hierarchical Sequential Cross-Attention over Chemical and Protein Language Models}

\author{%
\IEEEauthorblockN{%
Khadidja Henni\IEEEauthorrefmark{1},
Hamza Abdelali\IEEEauthorrefmark{3},
Abdelkrim Aries\IEEEauthorrefmark{3},\\
Neila Mezghani\IEEEauthorrefmark{1}\IEEEauthorrefmark{2},
Brigitte Vannier\IEEEauthorrefmark{4},
Sara Magdouli\IEEEauthorrefmark{5},
and Lina Abou-Abbas\IEEEauthorrefmark{6}}
\IEEEauthorblockA{\scriptsize
\IEEEauthorrefmark{1}I2A institute, T\'ELUQ University, Montreal, QC, Canada; \IEEEauthorrefmark{2}LIO, CRCHUM, Montreal, QC, Canada\\
\IEEEauthorrefmark{3}LCSI, ESI, Algiers, Algeria; \IEEEauthorrefmark{4}CoMeT UR 24344, Universite de Poitiers, Poitiers, France\\
\IEEEauthorrefmark{5}Dept. of Civil Engineering, University of Ottawa, Ottawa, ON, Canada; \IEEEauthorrefmark{6}Dept. of Electrical and Computer Engineering, Lebanese American University, Byblos, Lebanon}
}

\maketitle

\begin{abstract}
Predicting Drug-Target Interactions~(DTIs) is a central task in
computational drug discovery, with direct applications in virtual
screening, drug repurposing, and therapeutic candidate prioritization.
Although recent deep learning methods have improved DTI prediction,
many sequence-based models still process drugs and proteins independently
and only combine their representations at a late prediction stage.
This limits their ability to explicitly model cross-molecular dependencies
between chemical substructures and protein sequence regions.
In this paper, we propose a sequence-only DTI prediction architecture
that combines two pre-trained language models --- ChemBERTa for drug
SMILES strings and ESM-2 for protein amino acid sequences --- with a
hierarchical interaction module. The proposed model first extracts
contextual representations using pre-trained encoders, then applies
1D convolutional layers to condense local sequence patterns, followed
by a sequential bidirectional cross-attention mechanism inspired by
the induced-fit view of molecular recognition. Finally, attention-based
pooling constructs fixed-size interaction-aware vectors for binary
prediction. Experiments on BIOSNAP, Davis, and BindingDB show that the
proposed model achieves the best performance on BIOSNAP, matches the
best AUROC on Davis, and remains competitive on BindingDB while using
only \textbf{25.2 million trainable parameters}. Ablation results confirm
the contribution of both the CNN and cross-attention modules, and
cold-start experiments indicate promising generalization to unseen
proteins and drugs.
\end{abstract}

\begin{IEEEkeywords}
Drug-Target Interaction, Deep Learning, Pre-trained Language Models,
ChemBERTa, ESM-2, Cross-Attention, SMILES, Protein Sequences.
\end{IEEEkeywords}

\section{Introduction}

The development of new therapeutic compounds is a long, costly, and
high-risk process. From target identification to clinical approval,
drug discovery can require more than a decade of research and
substantial financial investment~\cite{hughes2011}. A key step in
this pipeline is the identification of Drug-Target Interactions~(DTIs),
which describe whether a chemical compound interacts with a biological
target, usually a protein. Accurate DTI prediction supports virtual
screening, drug repurposing, mechanism-of-action analysis, and early
safety assessment.

Traditionally, DTIs are identified through experimental screening.
Although \textit{in vitro} assays provide reliable evidence, they are
difficult to scale to the enormous number of possible drug-protein
pairs. The chemical space contains millions of possible drug-like
molecules, while the human proteome contains thousands of potential
protein targets. As a result, exhaustive experimental screening is
impractical. Computational approaches therefore play an important role
by prioritizing the most promising candidates for laboratory validation~\cite{hughes2011,lim2021}.

From a machine learning perspective, DTI prediction is often formulated
as a binary classification problem. Given a drug represented by a
SMILES string and a target represented by an amino acid sequence, the
model predicts whether an interaction exists. Early machine learning
methods relied heavily on handcrafted descriptors, molecular
fingerprints, and similarity-based features~\cite{lim2021}. Deep learning models
later reduced this dependence on manual feature engineering by learning
representations directly from raw molecular and biological sequences~\cite{deepdta}.

Several neural architectures have been explored for DTI prediction,
including convolutional neural networks~(CNNs)~\cite{deepdta},
recurrent neural networks~\cite{deepaffinity}, graph neural
networks~\cite{graphdta}, and Transformer-based
models~\cite{moltrans}. More recently, pre-trained language
models~(PLMs) have become increasingly important. ChemBERTa~\cite{chemberta}
learns chemical representations from large SMILES corpora, while
ESM-2~\cite{esm2} learns protein representations from large-scale
protein sequence databases. These models provide rich contextual
embeddings that can improve downstream DTI prediction.

Despite this progress, two limitations remain common in many
sequence-based models. First, drug and protein representations are
often computed independently and fused only at the final prediction
stage. This late-fusion strategy may fail to capture fine-grained
dependencies between drug substructures and protein sequence regions.
Second, global pooling or simple aggregation can discard
position-specific information that may be important for identifying
relevant residues or molecular fragments.

To address these limitations, we propose a hierarchical sequence-based
architecture for DTI prediction. The model combines ChemBERTa and
ESM-2 encoders with CNN-based local feature extraction, sequential
bidirectional cross-attention, and attention-based pooling. The
sequential cross-attention module models interaction-aware refinement
in two steps: the protein representation is first updated using drug
context, and the drug representation is then updated using the refined
protein context. This design is inspired by the induced-fit view of
molecular recognition~\cite{koshland}, without explicitly simulating
3D conformational changes.

The main contributions of this paper are as follows:
\begin{itemize}
    \item We propose a sequence-only DTI prediction architecture
    combining ChemBERTa and ESM-2 with CNN-based local feature
    extraction, achieving competitive performance without molecular
    graphs or structural data.
    \item We introduce a sequential bidirectional cross-attention
    mechanism that explicitly models asymmetric drug-protein
    dependencies and improves over its parallel counterpart on AUROC,
    AUPRC, and sensitivity, while achieving comparable specificity.
    \item We use attention-based pooling to construct
    interaction-aware drug and protein vectors while preserving
    position-level importance, enabling more focused aggregation
    than standard global pooling.
    \item Through systematic ablation, cold-start, and cross-dataset
    experiments, we demonstrate that each architectural component
    provides measurable improvements, and that the proposed partial
    fine-tuning strategy achieves competitive results with only 25.2M
    trainable parameters.
\end{itemize}

The rest of the paper is organized as follows.
Section~\ref{sec:related} presents related work.
Section~\ref{sec:method} describes the proposed architecture.
Section~\ref{sec:setup} details the experimental setup.
Section~\ref{sec:results} presents and discusses the results.
Section~\ref{sec:conclusion} concludes the paper and outlines
future work.

\section{Background and Related Work}
\label{sec:related}

\subsection{Drug and Protein Representations}

Drugs can be represented in several ways, including molecular
descriptors, fingerprints, molecular graphs, and sequence-based
notations. SMILES is one of the most widely used sequence
representations because it provides a compact textual encoding of
molecular structure~\cite{lim2021}. Although SMILES strings do not
explicitly describe 3D conformations, they are well suited for
sequence models and chemical language models.

Proteins are commonly represented by their amino acid sequences,
structural graphs, contact maps, or learned embeddings. Protein
sequences are widely available and can be processed by protein
language models trained on large biological databases. Such models
can capture contextual and evolutionary signals that are useful for
downstream prediction tasks.

\subsection{Deep Learning for DTI Prediction}

DeepDTA~\cite{deepdta} introduced a CNN-based approach that processes
SMILES and protein sequences directly. WideDTA~\cite{widedta} extended
this idea by using multiple sequence channels and motif-based
information. DeepAffinity~\cite{deepaffinity} used recurrent
architectures to model sequence dependencies, while
GraphDTA~\cite{graphdta} represented drugs as molecular graphs and
applied graph neural networks. More broadly, graph-based formulations
have also been used for high-dimensional representation, feature selection,
and clustering, illustrating the usefulness of graph structure for organizing
complex data beyond the DTI setting~\cite{henni2020clustergraph,henni2020isclustermpp}.

Transformer-based methods have also been applied to DTI prediction.
MolTrans~\cite{moltrans} uses self-attention to model substructure
interactions between drugs and proteins. Fine-tuning BERT-based models
has further shown that pre-trained sequence encoders can improve DTI
prediction performance~\cite{kang2022}. More recent hybrid methods
combine language models, graph encoders, and attention mechanisms,
often improving accuracy at the cost of higher architectural complexity
and larger trainable parameter counts~\cite{druglamp,dlmdti,henni2025structural}.

\subsection{Attention-Based Interaction Modeling}

Fusion strategies in DTI prediction can be broadly divided into late
fusion, self-attention over combined representations, and
cross-attention. Late fusion combines drug and protein vectors after
independent encoding, usually by concatenation. This strategy is
simple but may not fully capture cross-modal dependencies.
Cross-attention is more suitable for interaction modeling because one
modality can query the other, allowing the model to learn which drug
regions are relevant to which protein regions.

Most existing cross-attention approaches compute the interaction in a
single step or apply both attention directions simultaneously from the
original representations, without exploiting any sequential dependency
between the two refinement steps. This design gap motivates the
sequential hierarchical interaction module proposed in this work.

\section{Proposed Architecture}
\label{sec:method}

\subsection{Problem Formulation}

Let $\mathcal{D}$ be the set of drugs and $\mathcal{P}$ the set of
protein targets. Each drug $d \in \mathcal{D}$ is represented by a
SMILES string, and each protein $p \in \mathcal{P}$ by an amino acid
sequence. Given a drug-protein pair $(d, p)$, the objective is to
predict a binary label $y \in \{0, 1\}$, where $y = 1$ indicates an
interaction and $y = 0$ indicates no known interaction. The model
learns a prediction function:
\begin{equation}
\hat{y} = f\!\left(\mathrm{repr}(d),\, \mathrm{repr}(p)\right)
\in [0,1],
\label{eq:problem}
\end{equation}
where $\mathrm{repr}(\cdot)$ denotes the learned representation
produced by the corresponding encoder.

\subsection{Overall Architecture}

\begin{figure*}[t]
\centerline{\includegraphics[width=0.8\textwidth]{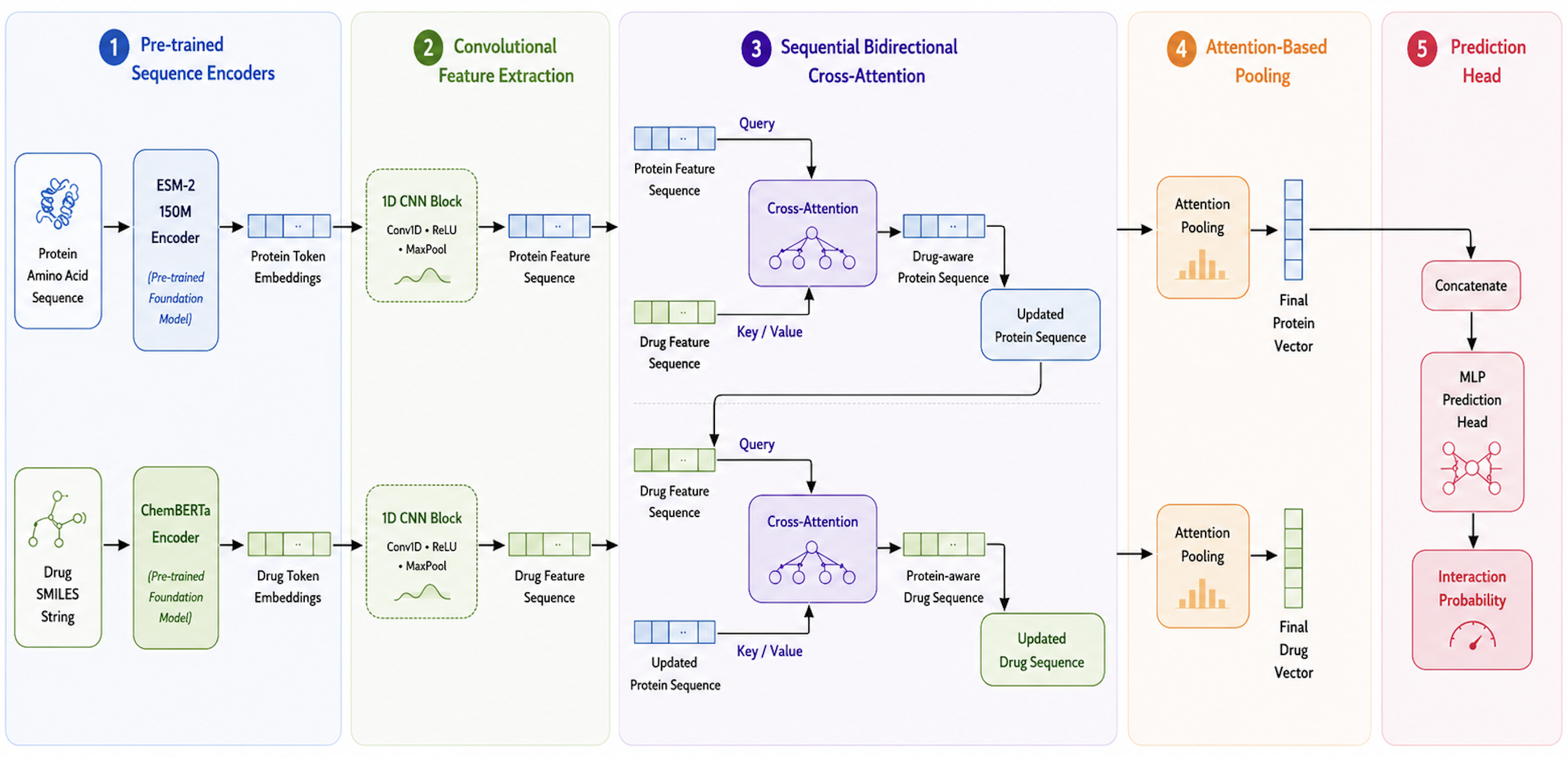}}
\caption{Overview of the proposed architecture. Drug SMILES and protein
sequences are encoded using ChemBERTa and ESM-2, condensed through 1D
CNN layers, fused using sequential bidirectional cross-attention, and
aggregated through attention-based pooling before final prediction.}
\label{fig:architecture}
\end{figure*}

Figure~\ref{fig:architecture} presents the proposed architecture,
which consists of five main stages: pre-trained sequence encoding,
CNN-based feature condensation, sequential bidirectional
cross-attention, and attention-based pooling followed by an MLP
prediction head.

\subsection{Pre-trained Sequence Encoders}

\paragraph{Protein Encoder (ESM-2):} For the protein branch, we use
\texttt{facebook/esm2\_t30\_150M\_UR50D}~\cite{esm2}, a 150M-parameter
protein language model pre-trained on the UniRef50 database via masked
language modeling~(MLM) over tens of millions of evolutionarily diverse
sequences. The encoder produces contextual amino acid embeddings of
dimension $d_p = 480$. The maximum protein sequence length is set to
1,024 tokens.

\paragraph{Drug Encoder (ChemBERTa):} For the drug branch, we use
\texttt{seyonec/ChemBERTa-zinc-base-v1}~\cite{chemberta,roberta}, a
RoBERTa-based chemical language model pre-trained on SMILES strings from
the ZINC database. It produces contextual drug token embeddings of
dimension $d_d = 768$. The maximum drug sequence length is set to
128 tokens.

Both encoders are initialized with pre-trained weights. To reduce
computational cost and limit overfitting, only the top two
Transformer layers of each encoder are unfrozen during fine-tuning;
all remaining layers are kept frozen. This partial fine-tuning strategy
reduces the number of trainable parameters from approximately 193
million to 25.2 million while preserving the knowledge acquired during
pre-training.

\subsection{CNN-Based Local Feature Extraction}

The contextual embeddings produced by the two encoders are passed
through independent 1D CNN blocks. Each block contains: (i)~a
\texttt{Conv1d} layer with kernel size~3 and 128 output channels; (ii)~a ReLU activation; and (iii)~a \texttt{MaxPool1d} layer with kernel size~2, reducing sequence length by half. The role of this stage is twofold: it captures local sequence patterns such as chemical fragments or protein motifs, and it reduces sequence length to make
subsequent cross-attention computation more efficient.

\subsection{Sequential Bidirectional Cross-Attention}

Let $\mathbf{H}_p \in \mathbb{R}^{L_p \times d}$ and $\mathbf{H}_d
\in \mathbb{R}^{L_d \times d}$ denote the CNN-condensed protein and
drug representations, where $L_p$, $L_d$ are the reduced sequence
lengths and $d = 128$ the shared hidden dimension after CNN projection.

Multi-Head Attention~(MHA) with $h$ heads is defined as:
\begin{equation}
\mathrm{MHA}(Q, K, V) =
\mathrm{Concat}(\mathrm{head}_1, \ldots, \mathrm{head}_h)\,W^O,
\label{eq:mha}
\end{equation}
where each head applies scaled dot-product attention:
\begin{equation}
\mathrm{Attention}(Q, K, V) =
\mathrm{softmax}\!\left(\frac{QK^\top}{\sqrt{d_k}}\right)V,
\label{eq:attention}
\end{equation}
with $h = 8$ heads and $d_k = d / h$.

The proposed interaction module proceeds in two {sequential}
steps:

\paragraph{Drug-Guided Protein Refinement} The protein
representation queries the drug representation, forcing its embeddings
to focus on the sub-regions most relevant to the drug's chemical
structure:
\begin{equation}
\mathbf{H}'_p = \mathrm{MHA}(Q = \mathbf{H}_p,\;
K = \mathbf{H}_d,\; V = \mathbf{H}_d).
\label{eq:protein_update}
\end{equation}

\paragraph{Adapted-Protein-Guided Drug Refinement} The
\textit{already drug-adapted} protein representation then refines the
drug features:
\begin{equation}
\mathbf{H}'_d = \mathrm{MHA}(Q = \mathbf{H}_d,\;
K = \mathbf{H}'_p,\; V = \mathbf{H}'_p).
\label{eq:drug_update}
\end{equation}

In a parallel design, both steps would use the original
representations, making them independent. Here, Step~2 uses the
refined $\mathbf{H}'_p$, creating a hierarchical dependency that
provides a richer joint representation. This sequential design is
inspired by the induced-fit principle~\cite{koshland}, without claiming
to simulate molecular dynamics or 3D conformational changes.

\subsection{Attention-Based Pooling and Prediction}

\paragraph{Attention Pooling} Standard global pooling discards
position-level information. We replace it with a learnable scoring
mechanism. For each position $i$, a scalar importance score is
computed as:
\begin{equation}
\alpha_i = \mathbf{w}_2^\top
\tanh\!\left(\mathbf{W}_1\,\mathbf{h}_i + \mathbf{b}_1\right) + b_2,
\label{eq:score}
\end{equation}
where $\mathbf{W}_1 \in \mathbb{R}^{d_a \times d}$,
$\mathbf{w}_2 \in \mathbb{R}^{d_a}$, and $d_a = 64$.
The scores are normalized and used to compute a weighted sum:
\begin{equation}
\bar{\alpha}_i = \frac{\exp(\alpha_i)}{\sum_j \exp(\alpha_j)},
\qquad
\mathbf{z} = \sum_{i=1}^{L} \bar{\alpha}_i\,\mathbf{h}_i.
\label{eq:pooling}
\end{equation}
This concentrates the fixed-size representation on pharmacologically
relevant positions while suppressing uninformative regions.

\paragraph{Prediction Head} The pooled vectors $\mathbf{z}_p$ and
$\mathbf{z}_d$ are concatenated to form a vector of dimension $2d$
and passed to an MLP with three linear layers
($2d \to 256 \to 128 \to 1$), ReLU activations, and
Dropout~($p = 0.2$) after each hidden layer. The final logit is mapped
to a probability via sigmoid. Training minimizes the binary
cross-entropy with logits loss:
\begin{equation}
\mathcal{L} =
-\frac{1}{N}\sum_{i=1}^{N}\!\left[
y_i \log\sigma(\hat{y}_i)
+ (1 - y_i)\log(1 - \sigma(\hat{y}_i))
\right],
\label{eq:loss}
\end{equation}
where $\sigma$ denotes the sigmoid function.

\section{Experimental Setup}
\label{sec:setup}

\subsection{Datasets}

The proposed model is evaluated on three benchmark datasets:
Davis, BIOSNAP, and BindingDB. Their statistics are summarized
in Table~\ref{tab:datasets}.

\paragraph{Davis}~\cite{davis2011} contains binding affinity measurements
between kinase inhibitors and kinase proteins. Following common binary
DTI practice, affinities are binarized using a threshold of
$K_d < 30$~nM, yielding a severely imbalanced dataset
($\approx$1:6 ratio).

\paragraph{BIOSNAP}~\cite{biosnap} is a balanced drug-target interaction
benchmark derived from the BioSNAP biomedical network collection. It
contains validated positive interactions and sampled negative pairs in
equal proportion, making it suitable for evaluating discriminative
performance without imbalance bias.

\paragraph{BindingDB}~\cite{bindingdb} is a large public database of
experimentally measured protein-ligand binding affinities. Binary
labels are derived using a standard threshold of $K_d < 1\,\mu\mathrm{M}$.

For Davis and BindingDB, the training sets are resampled to class
balance to mitigate the effect of imbalance during optimization.
Generative augmentation of latent drug--target representations has also
been investigated as an alternative strategy for DTI class imbalance~\cite{silva2026matrix}.
Validation and test sets retain their original distributions to
reflect realistic evaluation conditions. No additional negative
sampling is introduced during testing.

\begin{table}[htbp]
\caption{Dataset statistics. Interaction pairs are reported as
train\,/\,validation\,/\,test.}
\centering
\renewcommand{\arraystretch}{1.15}
\setlength{\tabcolsep}{2.5pt}
\begin{tabular}{lcccc}
\toprule
\textbf{Dataset} & \textbf{Drugs} & \textbf{Proteins}
& \textbf{Positive Pairs} & \textbf{Negative Pairs} \\
\midrule
Davis     & 68     & 379   & 1,043\,/\,160\,/\,303
                           & 1,043\,/\,2,846\,/\,5,708 \\
BIOSNAP   & 4,510  & 2,181 & 9,619\,/\,1,374\,/\,2,748
                           & 9,619\,/\,1,374\,/\,2,748 \\
BindingDB & 10,665 & 1,413 & 6,334\,/\,927\,/\,1,905
                           & 6,334\,/\,5,717\,/\,11,384 \\
\bottomrule
\end{tabular}
\label{tab:datasets}
\end{table}

\subsection{Evaluation Protocol}

We use two evaluation settings. In the {standard pair-level}
setting, drug-protein pairs are divided into training, validation,
and test sets. A drug or protein may appear across multiple splits
with different partners, measuring the model's ability to predict
new interaction pairs for known entities.

In the {cold-start} setting, 20\% of drugs (resp.\ proteins)
are entirely held out, and all their interactions form the test set.
This evaluates generalization to molecular entities not observed
during training. Recent novelty-aware DTI work further emphasizes
separate evaluation of ligand and protein novelty, since the two
settings can induce different generalization challenges~\cite{abouabbas2026novelty}.

During development, all proposed-model experiments were repeated over
five independent runs. The reported values correspond to the test
performance of the checkpoint selected according to validation AUROC.
Since not all baseline publications report standard deviations under
identical splits, the comparison table reports point estimates, and
small differences should be interpreted cautiously. A fully controlled
benchmark with mean and standard deviation over all seeds is left for
future work.

\subsection{Training Configuration}

The model is implemented using PyTorch and PyTorch Lightning.
Key hyperparameters are reported in Table~\ref{tab:hparams}. Early
stopping monitors validation AUROC with a patience of three epochs.

\begin{table}[htbp]
\caption{Main hyperparameters of the proposed model.}
\centering
\renewcommand{\arraystretch}{1.15}
\begin{tabular}{lc}
\toprule
\textbf{Hyperparameter} & \textbf{Value} \\
\midrule
Optimizer                  & AdamW \\
Base learning rate         & $5 \times 10^{-5}$ \\
Learning-rate schedule     & Linear warmup + cosine decay \\
Effective batch size       & 32 \\
Maximum epochs             & 30 \\
Unfrozen encoder layers    & Top 2 layers \\
CNN output channels        & 128 \\
Attention pooling dim $d_a$ & 64 \\
MLP hidden layers          & $2d \to 256 \to 128 \to 1$ \\
MLP dropout                & 0.2 \\
Protein max length         & 1,024 tokens \\
Drug max length            & 128 tokens \\
\bottomrule
\end{tabular}
\label{tab:hparams}
\end{table}

\subsection{Metrics}

The model is evaluated using four complementary metrics: AUROC
measures the global ability to separate positive and negative pairs
across all classification thresholds; AUPRC is especially relevant
for imbalanced datasets as it focuses on positive class performance;
Sensitivity measures the proportion of true interactions correctly
detected; and Specificity measures the proportion of non-interactions
correctly rejected.

\section{Results and Discussion}
\label{sec:results}

\subsection{Comparison with State-of-the-Art Methods}

We compare the proposed model against representative DTI prediction
methods spanning multiple architectural paradigms: CNN-based
(DeepDTA~\cite{deepdta}), Transformer-based
(MolTrans~\cite{moltrans}), PLM fine-tuning
(Fine-tuning BERT~\cite{kang2022}), dual PLM
(DLM-DTI~\cite{dlmdti}), interpretable cross-attention
(ICAN~\cite{ican}), HyperAttentionDTI~\cite{hyperatt}, and a
recent multi-scale fusion approach (Yang et al.~\cite{yang2024}).
Baseline values are taken from the corresponding publications or
comparable reported benchmark results when available. The comparison
should therefore be interpreted as a literature-level evaluation
rather than a fully controlled reimplementation of all baselines.

Table~\ref{tab:main_results} presents results on all three datasets.

\paragraph{BIOSNAP.} The proposed model obtains the best AUROC~(0.920)
and AUPRC~(0.924) among all compared methods, demonstrating that
the combination of pre-trained encoders, CNN feature extraction, and
sequential cross-attention is effective on balanced DTI data. The
model achieves a sensitivity of 0.854 and specificity of 0.830,
slightly below Fine-tuning BERT on these two metrics, but leading
overall.

\paragraph{Davis.} The proposed model reaches an AUROC of 0.920,
matching the best reported result. It also obtains the highest
sensitivity~(0.901), which is important in virtual screening because
false negatives represent missed candidate interactions. However, its
AUPRC~(0.356) is the lowest in the comparison on this dataset,
indicating that precision under severe class imbalance remains a clear
limitation of the current approach.

\paragraph{BindingDB.} On this large-scale, chemically diverse dataset,
the proposed model achieves a competitive AUROC of 0.920 and
AUPRC of 0.642. It does not lead on every metric --- HyperAttentionDTI
achieves the best AUPRC~(0.876) and DLM-DTI achieves the best
specificity~(0.916) --- but it remains competitive while relying
exclusively on sequence inputs without molecular graphs or 3D
structural data.

\begin{table*}[t]
\caption{Performance comparison on BIOSNAP, Davis, and BindingDB.
Best result per metric per dataset is in \textbf{bold}.}
\centering
\renewcommand{\arraystretch}{1.15}
\setlength{\tabcolsep}{3.5pt}
\begin{tabular}{lcccccccccccc}
\toprule
 & \multicolumn{4}{c}{\textbf{BIOSNAP}}
 & \multicolumn{4}{c}{\textbf{Davis}}
 & \multicolumn{4}{c}{\textbf{BindingDB}} \\
\cmidrule(lr){2-5}\cmidrule(lr){6-9}\cmidrule(lr){10-13}
\textbf{Model}
 & \textbf{AUROC} & \textbf{AUPRC} & \textbf{Sens.} & \textbf{Spec.}
 & \textbf{AUROC} & \textbf{AUPRC} & \textbf{Sens.} & \textbf{Spec.}
 & \textbf{AUROC} & \textbf{AUPRC} & \textbf{Sens.} & \textbf{Spec.} \\
\midrule
MolTrans~\cite{moltrans}
 & 0.895 & 0.901 & 0.775 & 0.851
 & 0.907 & 0.404 & 0.800 & 0.876
 & 0.914 & 0.622 & 0.797 & 0.896 \\
DeepDTA~\cite{deepdta}
 & 0.834 & 0.849 & 0.726 & 0.813
 & 0.792 & 0.272 & 0.688 & 0.778
 & 0.902 & 0.604 & 0.757 & 0.904 \\
HyperAttentionDTI~\cite{hyperatt}
 & 0.825 & 0.830 & 0.800 & 0.812
 & 0.841 & 0.483 & 0.708 & 0.722
 & 0.900 & \textbf{0.876} & 0.801 & 0.791 \\
Fine-tuning BERT~\cite{kang2022}
 & 0.914 & 0.900 & \textbf{0.862} & \textbf{0.847}
 & \textbf{0.920} & 0.395 & 0.824 & 0.802
 & \textbf{0.922} & 0.623 & 0.814 & 0.793 \\
DLM-DTI~\cite{dlmdti}
 & 0.914 & 0.914 & 0.848 & 0.844
 & 0.895 & 0.373 & 0.833 & 0.766
 & 0.912 & 0.643 & 0.846 & \textbf{0.916} \\
ICAN~\cite{ican}
 & 0.871 & 0.886 & 0.799 & 0.786
 & 0.903 & 0.372 & 0.884 & 0.765
 & 0.900 & 0.604 & \textbf{0.857} & 0.815 \\
Yang et al.~\cite{yang2024}
 & 0.885 & 0.856 & 0.828 & 0.787
 & 0.799 & \textbf{0.515} & 0.896 & 0.850
 & 0.910 & 0.659 & 0.815 & 0.818 \\
\midrule
\textbf{Proposed}
 & \textbf{0.920} & \textbf{0.924} & 0.854 & 0.830
 & \textbf{0.920} & 0.356 & \textbf{0.901} & 0.801
 & 0.920 & 0.642 & 0.816 & 0.874 \\
\bottomrule
\end{tabular}
\label{tab:main_results}
\end{table*}

\subsection{Integrated Dataset Evaluation}

To evaluate whether the model benefits from a larger and more diverse
biochemical training corpus, we train it on the combined training and
validation sets of Davis, BIOSNAP, and BindingDB, then evaluate
separately on each original test set.

As shown in Table~\ref{tab:integrated}, integrated training yields
substantial AUPRC improvements on Davis~($+0.092$: $0.356 \to 0.448$)
and BindingDB~($+0.122$: $0.642 \to 0.764$), confirming that the
architecture scales well with data diversity. On BIOSNAP, AUPRC
decreases from 0.924 to 0.840, which may reflect domain shift caused
by mixing datasets with different distributions and label-generation
protocols.

\begin{table}[htbp]
\caption{Individual training vs.\ integrated training.}
\centering
\renewcommand{\arraystretch}{1.15}
\setlength{\tabcolsep}{3pt}
\begin{tabular}{llcccc}
\toprule
\textbf{Test set} & \textbf{Training}
 & \textbf{AUROC} & \textbf{AUPRC} & \textbf{Sens.} & \textbf{Spec.} \\
\midrule
\multirow{2}{*}{Davis}
 & Individual & 0.920 & 0.356 & 0.901 & 0.801 \\
 & Integrated & \textbf{0.939} & \textbf{0.448}
              & \textbf{0.911} & \textbf{0.848} \\
\midrule
\multirow{2}{*}{BIOSNAP}
 & Individual & \textbf{0.920} & \textbf{0.924} & 0.854 & 0.830 \\
 & Integrated & 0.920 & 0.840 & \textbf{0.886} & \textbf{0.835} \\
\midrule
\multirow{2}{*}{BindingDB}
 & Individual & \textbf{0.920} & 0.642 & 0.816 & \textbf{0.874} \\
 & Integrated & 0.918 & \textbf{0.764} & \textbf{0.852} & 0.850 \\
\bottomrule
\end{tabular}
\label{tab:integrated}
\end{table}

\subsection{Cold-Start Generalization}

To assess generalization to unseen entities, we conduct a cold-start
evaluation using a 20\% hold-out protocol applied independently to
drugs and proteins. Results are reported in Table~\ref{tab:coldstart}.

In the {unseen-protein} setting, the proposed model achieves
a ROC-AUC of 0.87, outperforming MolTrans~(0.77) and
DeepDTA~(0.85). This suggests that ESM-2 transfers useful
evolutionary and structural priors that help the model reason about
novel protein targets beyond its training distribution. In the
{unseen-drug} setting, all three models reach 0.85, indicating
that sequence-based approaches generalize comparably to new chemical
entities in this evaluation.

\begin{table}[htbp]
\caption{Cold-start evaluation (ROC-AUC).}
\centering
\renewcommand{\arraystretch}{1.15}
\begin{tabular}{lcc}
\toprule
\textbf{Model} & \textbf{Unseen Drugs} & \textbf{Unseen Proteins} \\
\midrule
MolTrans~\cite{moltrans}  & 0.85 & 0.77 \\
DeepDTA~\cite{deepdta}    & 0.85 & 0.85 \\
\textbf{Proposed}         & 0.85 & \textbf{0.87} \\
\bottomrule
\end{tabular}
\label{tab:coldstart}
\end{table}

\subsection{Ablation Study}

We conduct an ablation study on BIOSNAP to evaluate the contribution
of each architectural component. Results are shown in
Table~\ref{tab:ablation}.

\paragraph{Component importance}
Removing the CNN block reduces AUROC from 0.920 to 0.916, confirming
that local feature extraction provides complementary information
beyond the Transformer embeddings alone. Removing the cross-attention
block causes a larger degradation~(AUROC: $0.920 \to 0.911$),
confirming that cross-modal interaction modeling is the single most
important post-encoder component. Removing both modules yields the
weakest performance~(AUROC: 0.905), validating their complementary
and cumulative contributions.

\paragraph{Sequential vs.\ parallel attention:}
The sequential variant outperforms the parallel variant on
AUROC~(0.920 vs.\ 0.914), AUPRC~(0.924 vs.\ 0.920), and
Sensitivity~(0.854 vs.\ 0.833), while the parallel variant slightly
improves Specificity~(0.834 vs.\ 0.830). The gain is moderate but
consistent across the main discriminative metrics, supporting the
hypothesis that hierarchical refinement --- where Step~2 queries an
already drug-adapted protein context --- provides a richer joint
representation than computing both directions simultaneously.

\begin{table}[htbp]
\caption{Ablation study on BIOSNAP.}
\centering
\renewcommand{\arraystretch}{1.15}
\begin{tabular}{lcccc}
\toprule
\textbf{Variant}
 & \textbf{AUROC} & \textbf{AUPRC} & \textbf{Sens.} & \textbf{Spec.} \\
\midrule
Full model              & \textbf{0.920} & \textbf{0.924}
                        & \textbf{0.854} & 0.830 \\
Without CNN             & 0.916 & 0.920 & 0.849 & 0.828 \\
Without attention       & 0.911 & 0.914 & 0.835 & 0.819 \\
Without CNN + attention & 0.905 & 0.907 & 0.828 & 0.812 \\
\midrule
Sequential attention    & \textbf{0.920} & \textbf{0.924}
                        & \textbf{0.854} & 0.830 \\
Parallel attention      & 0.914 & 0.920 & 0.833 & \textbf{0.834} \\
\bottomrule
\end{tabular}
\label{tab:ablation}
\end{table}

\subsection{Model Complexity}

By restricting fine-tuning to the top two layers of each encoder and
the newly added modules, the proposed architecture achieves only
{25.2 million trainable parameters}, despite a total model
size of approximately 193 million. As shown in
Table~\ref{tab:complexity}, this represents a 2.5$\times$ reduction
relative to MolTrans, a 3.4$\times$ reduction relative to DLM-DTI,
and a 14$\times$ reduction relative to full BERT fine-tuning, while
maintaining competitive or superior predictive performance.

\begin{table}[htbp]
\caption{Trainable parameter comparison.}
\centering
\renewcommand{\arraystretch}{1.15}
\begin{tabular}{lc}
\toprule
\textbf{Model} & \textbf{Trainable Parameters} \\
\midrule
MolTrans~\cite{moltrans}         & 62.8M  \\
DLM-DTI~\cite{dlmdti}            & 86.7M  \\
Fine-tuning BERT~\cite{kang2022} & 353.0M \\
\textbf{Proposed}                & \textbf{25.2M} \\
\bottomrule
\end{tabular}
\label{tab:complexity}
\end{table}

\subsection{Discussion and Limitations}

The proposed architecture performs best on BIOSNAP, where the balanced
class distribution allows both AUROC and AUPRC to reflect genuine
discriminative quality. On Davis, high AUROC and sensitivity are
achieved, but the low AUPRC~(0.356) reveals a precision limitation
under severe class imbalance: the model prioritizes recall at the
expense of precision, which may be acceptable in early-stage virtual
screening but would require recalibration for applications demanding
higher specificity. On BindingDB, competitive AUROC is obtained
despite the chemical diversity of the dataset, suggesting that
ChemBERTa provides sufficiently general representations. However, the
gap in AUPRC relative to HyperAttentionDTI, which uses structural
information, points to a ceiling for purely sequence-based approaches
on chemically complex data.

Several limitations should be noted. First, the method is
sequence-only and does not use molecular graphs, 3D protein
structures, docking poses, or binding-site geometry, which may limit
performance on datasets where structural information is decisive.
Second, although attention weights provide a more transparent
interaction mechanism than simple concatenation, they should not be
interpreted as biological explanations without validation against
known binding sites or crystal structures. Third, the baseline
comparison relies partly on reported literature values obtained under
potentially different preprocessing and split protocols; a fully
controlled reimplementation would be required for strict comparison.

\section{Conclusion }
\label{sec:conclusion}

This paper presented a hierarchical sequence-based architecture for
Drug-Target Interaction prediction combining ChemBERTa and ESM-2
pre-trained language models with CNN-based local feature extraction,
sequential bidirectional cross-attention, and attention-based pooling.
Experiments on BIOSNAP, Davis, and BindingDB demonstrate that the
proposed method achieves strong and competitive results while using
only 25.2 million trainable parameters. Ablation experiments confirm
that both the CNN condensation and the sequential cross-attention
module provide measurable improvements over their respective
ablated variants. Cold-start evaluation indicates encouraging
generalization, particularly to unseen protein targets. The Discussion
section explicitly identifies precision under class imbalance and the
absence of structural information as current limitations.

Future work will focus on four directions. First, richer chemical encoders or graph-based drug representations could be integrated to capture molecular topology more explicitly, which may reduce the AUPRC
gap on imbalanced and structurally diverse datasets. Second, protein
structural information or predicted binding-site features could be
added to provide better biological grounding. Third, attention maps
should be systematically validated against known binding residues or
protein-ligand crystal structures before making strong interpretability claims. Finally, a fully controlled benchmark with reimplemented baselines, identical splits, and reported mean and standard deviation
across multiple seeds would provide a more rigorous empirical foundation for future work in this line of research.

\section*{Generative AI Usage Statement}

Generative AI tools were used only for language polishing and formatting
assistance. All AI-assisted revisions were reviewed and verified by the
authors. All scientific contributions, experiments, results, and
conclusions are entirely produced by  the authors.

\end{document}